\documentclass[conference]{IEEEtran}
\IEEEoverridecommandlockouts
\usepackage{cite}
\usepackage{amsmath,amssymb,amsfonts}
\usepackage{algorithmic}
\usepackage{graphicx}
\usepackage{textcomp}
\usepackage{xcolor}
\usepackage{subfigure}
\usepackage{float}
\usepackage{subcaption}
\usepackage[autostyle=true]{csquotes}

\def\BibTeX{{\rm B\kern-.05em{\sc i\kern-.025em b}\kern-.08em
    T\kern-.1667em\lower.7ex\hbox{E}\kern-.125emX}}
\begin{document}

\title{Gen-AI Police Sketches with Stable Diffusion}

\author{
\IEEEauthorblockN{Aaron Contreras}
\IEEEauthorblockA{\textit{Computer Science} \\
\textit{Harvard College}\\
Cambridge, MA \\
aaroncontreras@college.\\
harvard.edu}
\and
\IEEEauthorblockN{Nico Fidalgo}
\IEEEauthorblockA{\textit{Computer Science} \\
\textit{Harvard College}\\
Cambridge, MA \\
nfidalgo@college.\\
harvard.edu}
\and
\IEEEauthorblockN{Katherine Harvey}
\IEEEauthorblockA{\textit{Computer Science} \\
\textit{Harvard College}\\
Cambridge, MA \\
kharvey@college.\\
harvard.edu}
\and
\IEEEauthorblockN{Johnny Ni}
\IEEEauthorblockA{\textit{Computer Science} \\
\textit{Harvard College}\\
Cambridge, MA \\
johnnyni@college.\\
harvard.edu}
}

\maketitle

\begin{abstract}
This project investigates the use of multimodal AI-driven approaches to automate and enhance suspect sketching. Three pipelines were developed and evaluated: (1) baseline image-to-image Stable Diffusion model, (2) same model integrated with a pre-trained CLIP model for text-image alignment, and (3) novel approach incorporating LoRA fine-tuning of the CLIP model, applied to self-attention and cross-attention layers, and integrated with Stable Diffusion. An ablation study confirmed that fine-tuning both self- and cross-attention layers yielded the best alignment between text descriptions and sketches. Performance testing revealed that Model 1 achieved the highest structural similarity (SSIM) of 0.72 and a peak signal-to-noise ratio (PSNR) of 25 dB, outperforming Model 2 and Model 3. Iterative refinement enhanced perceptual similarity (LPIPS), with Model 3 showing improvement over Model 2 but still trailing Model 1. Qualitatively, sketches generated by Model 1 demonstrated the clearest facial features, highlighting its robustness as a baseline despite its simplicity.
\end{abstract}

\section{Introduction}

Generating accurate police sketches is a critical task in law enforcement when photographs are unavailable, requiring effective visual representation of suspects. Traditional methods rely on manual artistry, which is time-consuming and inconsistent. Recent advancements in AI present an opportunity to automate and enhance this process, making it more efficient and reliable.

\begin{figure}[h!]
    \centering
    \includegraphics[width=0.5\textwidth]{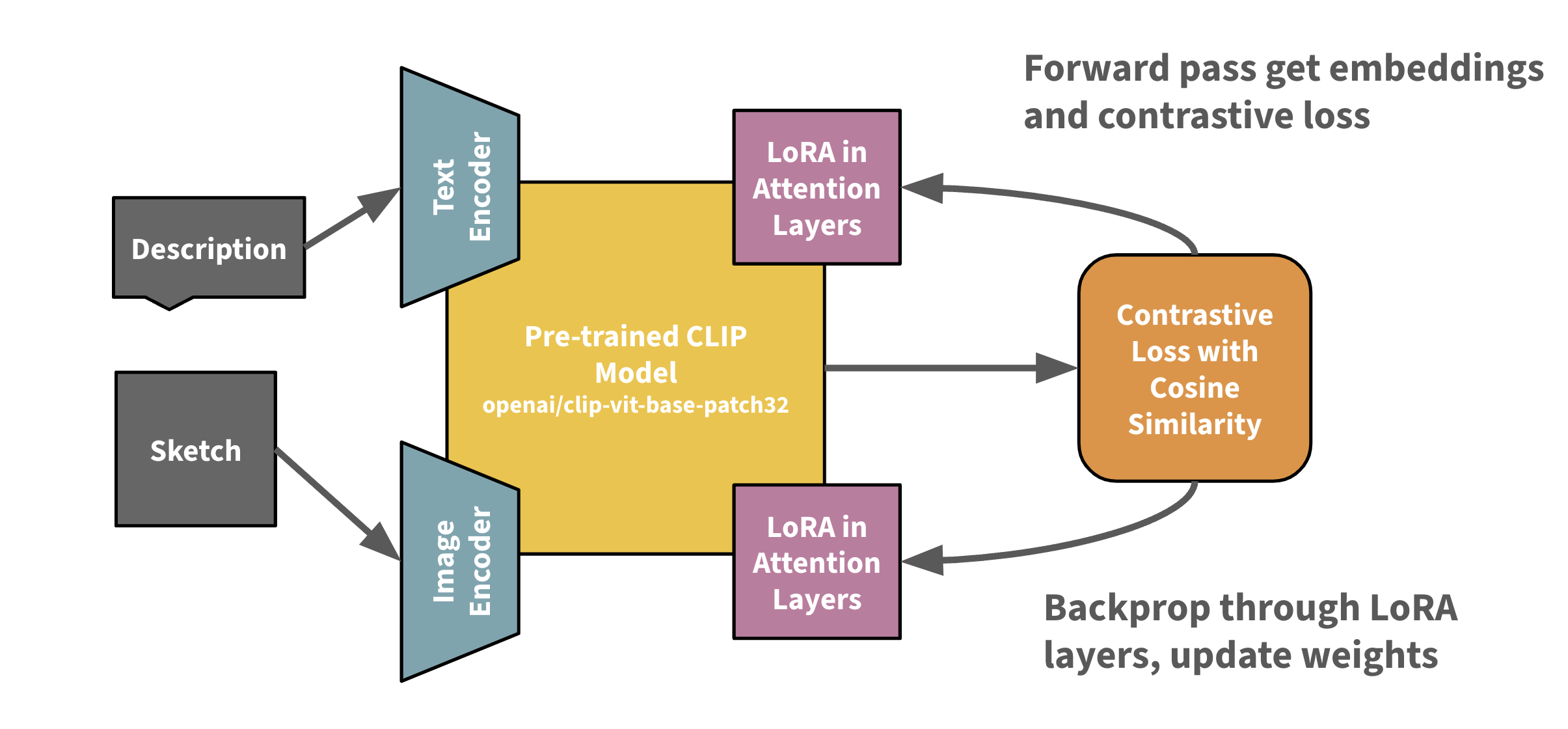}
    \caption{Pipeline of our novel approach (Model 3). The CLIP model is fine-tuned using LoRA applied to attention layers, optimizing contrastive loss to improve alignment between text descriptions and sketches.}
    \label{fig:proposed_approach}
\end{figure}

This project investigates multimodal AI models to generate police sketches from text descriptions and initial sketches. We evaluated three approaches: (1) a baseline Stable Diffusion model, (2) the same model integrated with a pre-trained CLIP model for enhanced text-image alignment, and (3) a novel approach using LoRA-based fine-tuning of CLIP integrated with Stable Diffusion. Figure \ref{fig:proposed_approach} illustrates the pipeline of Model (3).

Our approach addresses limitations of traditional methods by enabling iterative refinement, allowing users to dynamically improve sketches. This capability enhances usability and accuracy, streamlining the sketch generation process and improving investigative outcomes. Iterative refinement steps include embedding updates and prompt adjustments, which improve image quality over successive iterations.

\section{Approach}

We evaluated three Stable Diffusion models tasked with generating police sketches using multimodal inputs:

\begin{enumerate}
    \item \textbf{Baseline Stable Diffusion Model:} This model (\texttt{runwayml/stable-diffusion-v1-5}) generates sketches directly from input sketches.
    \item \textbf{Stable Diffusion with Pre-trained CLIP:} This model integrates a pre-trained CLIP model (\texttt{openai/clip-vit-base-patch32}) to improve text-image alignment in sketch generation.
    \item \textbf{Novel Approach with Fine-tuned CLIP:} Using LoRA, we fine-tuned the CLIP model to improve its ability to capture nuanced relationships between text descriptions and sketches. The fine-tuned model was integrated into the Stable Diffusion pipeline.
\end{enumerate}

Our key novelty lies in fine-tuning both self-attention and cross-attention layers of the CLIP model using LoRA, enhancing alignment and image quality. Additionally, iterative refinement was applied to all models to optimize outputs based on user feedback. This process combines updated embeddings from text and image inputs, enabling dynamic adjustments and improved sketches.

\section{Implementation}

\subsection{Dataset Preparation}
We curated a dataset with 295 (description, sketch) pairs from the CUHK Face Sketch FERET Database (CUFSF). To ensure uniformity, structured text descriptions were generated using ChatGPT-4, following a template format (e.g., \enquote{The suspect is described as [demographic] with [physical attributes]...}). This dataset facilitated effective fine-tuning of the CLIP model for text-sketch alignment.

\subsection{Model Pipeline}

\begin{itemize}
    \item \textbf{Baseline and Pre-trained CLIP Models:} We integrated \texttt{runwayml/stable-diffusion-v1-5} and \texttt{openai/clip-vit-base-patch32} models using the Hugging Face library, without additional training.
    \item \textbf{Fine-tuned CLIP Model:} LoRA-based fine-tuning was applied to self- and cross-attention layers of the CLIP model, as guided by ablation studies, before integration with the Stable Diffusion pipeline.
\end{itemize}

\subsection{Iterative Refinement Process}

\begin{itemize}
    \item Text prompts were truncated to 77 tokens to match CLIP's token limit.
    \item Combined embeddings from the input image and updated prompts were generated, normalized, and projected to match Stable Diffusion's input requirements.
    \item The input image was preprocessed into latents using Stable Diffusion's VAE encoder, ensuring compatibility.
    \item Stable Diffusion generated an output image guided by the embeddings, with strength (0.3) controlling image deviation and guidance scale (7.5) determining prompt influence.
    \item Generated images were re-encoded into updated latents for cumulative refinements, allowing alignment with evolving prompts over iterations.
\end{itemize}

\subsection{Fine-tuning Process}
Fine-tuning was performed on an 80/20 train/validation split of the dataset. Training loss and top-k accuracy metrics were tracked over epochs to monitor performance. Top-k accuracy measured the likelihood of correctly identifying the corresponding image from its text within top-1, top-5, top-10, and top-25 predictions.

\subsection{Performance Metrics}

\begin{itemize}
    \item \textbf{Structural Similarity Index (SSIM):} Measures structural alignment between images, with higher values indicating better similarity.
    \item \textbf{Peak Signal-to-Noise Ratio (PSNR):} Quantifies pixel-level clarity, where higher values indicate reduced distortion.
    \item \textbf{CLIP Score:} Assesses similarity between generated images and text descriptions using the CLIP model.
    \item \textbf{Learned Perceptual Image Patch Similarity (LPIPS):} Measures perceptual similarity, with lower values indicating closer resemblance.
\end{itemize}

\section{Results}

\begin{figure}[h!]
    \centering
    \includegraphics[width=0.15\textwidth]{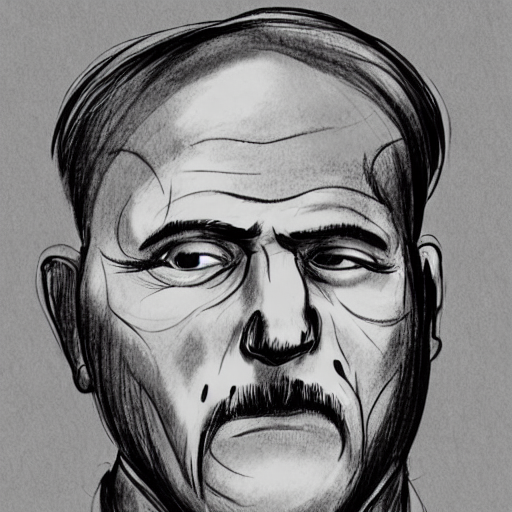}
    \includegraphics[width=0.15\textwidth]{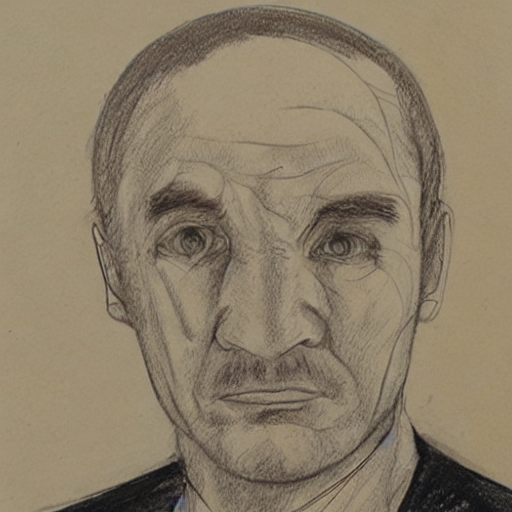}
    \includegraphics[width=0.15\textwidth]{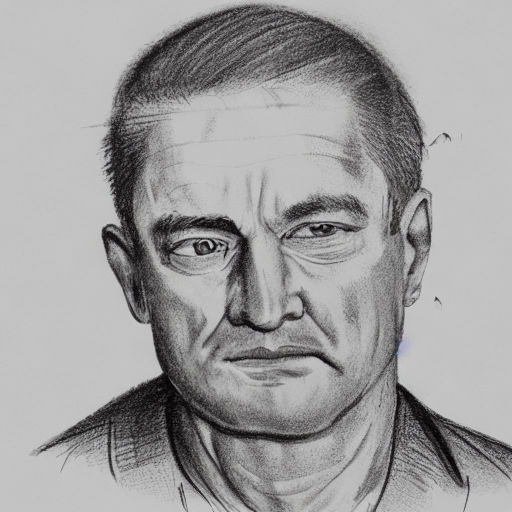}
    \caption{Results of fine-tuning on self-attention layers only (left), cross-attention layers only (center), and both layers (right). Fine-tuning both layers resulted in better alignment and image quality, as observed in the refined facial features and overall structure.}
    \label{fig:ablation_study}
\end{figure}

To determine which layers of the CLIP model to fine-tune, we conducted an ablation study. Figure \ref{fig:ablation_study} shows the results of fine-tuning on self-attention layers only, cross-attention layers only, and both. The results indicate that targeting both self- and cross-attention layers provides the best visual quality, as combining these layers allows the model to better capture relationships between text and image features.

\begin{figure}[h!]
    \centering
    \includegraphics[width=0.5\textwidth]{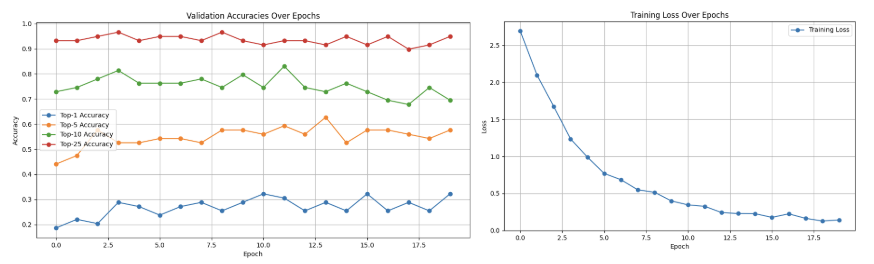}
    \caption{Left: Validation accuracy (Top-1, Top-5, Top-10, and Top-25) over epochs. Right: Training loss over epochs. The training loss curve shows steady convergence, indicating effective fine-tuning of the model. Validation accuracy trends highlight the model's improving ability to align text and sketches, with Top-25 accuracy reaching above 90\%.}
    \label{fig:fine_tuning_metrics}
\end{figure}

The training loss when fine-tuning the CLIP model for Model (3) exhibited a consistent decay as shown in Figure \ref{fig:fine_tuning_metrics}, approaching an asymptote around epoch 15, suggesting convergence. The validation accuracy curves demonstrate steady improvement indicating that the model successfully aligned multimodal inputs effectively.

\begin{figure}[h!]
    \centering
    \includegraphics[width=0.5\textwidth]{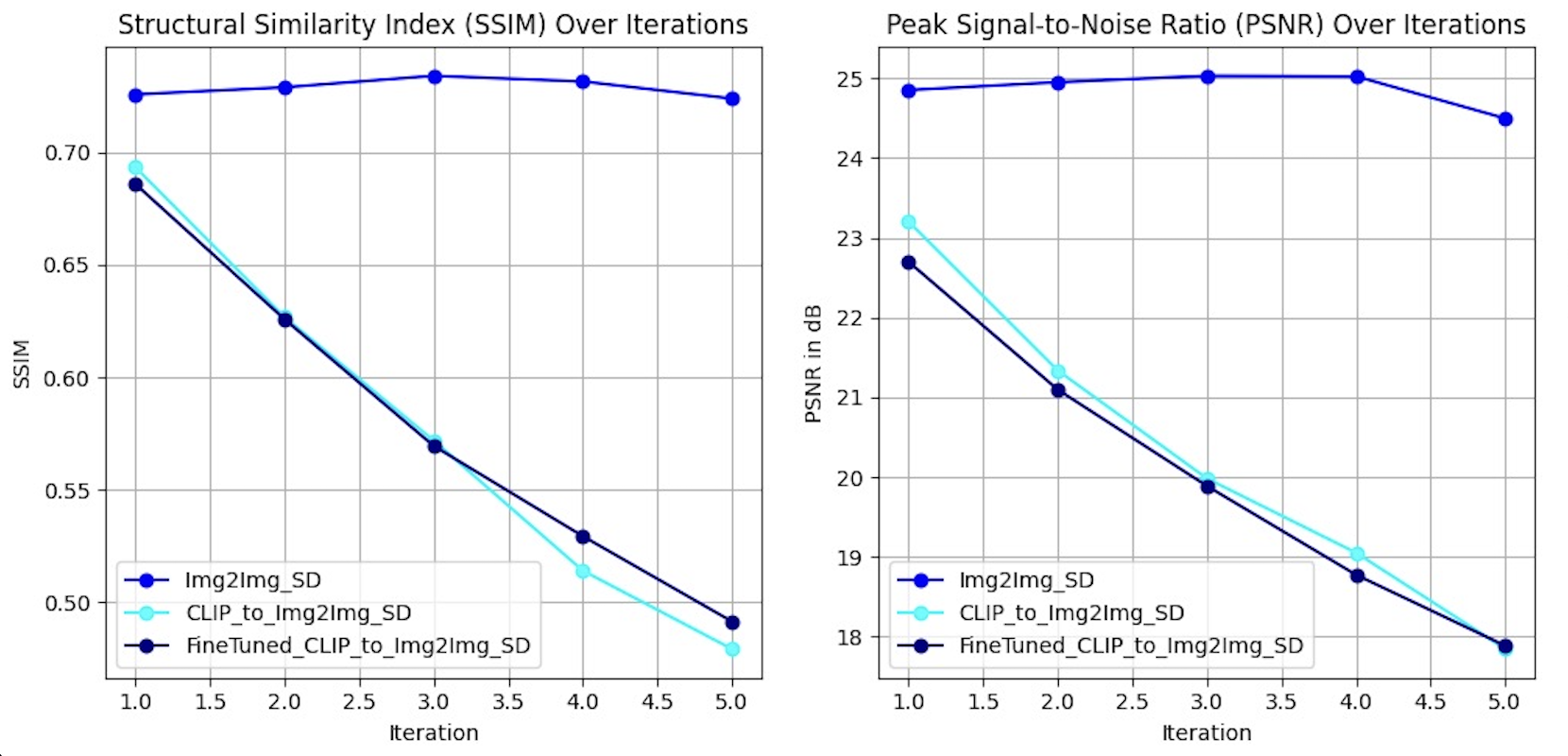}
    \caption{Left: SSIM over iterative refinement. Right: PSNR over iterative refinement. Model (1) consistently has the highest SSIM performing 20.84\% better than Model (2) and 19.72\% better than Model (3). It also has the highest PSNR performing 20.95\% better than Model (2) and 25\% better than Model (3) across all iterations, suggesting more consistency between iterations and reduced distortion compared to Model (2) and Model (3).}
    \label{fig:tests1}
\end{figure}

\begin{figure}[h!]
    \centering
    \includegraphics[width=0.5\textwidth]{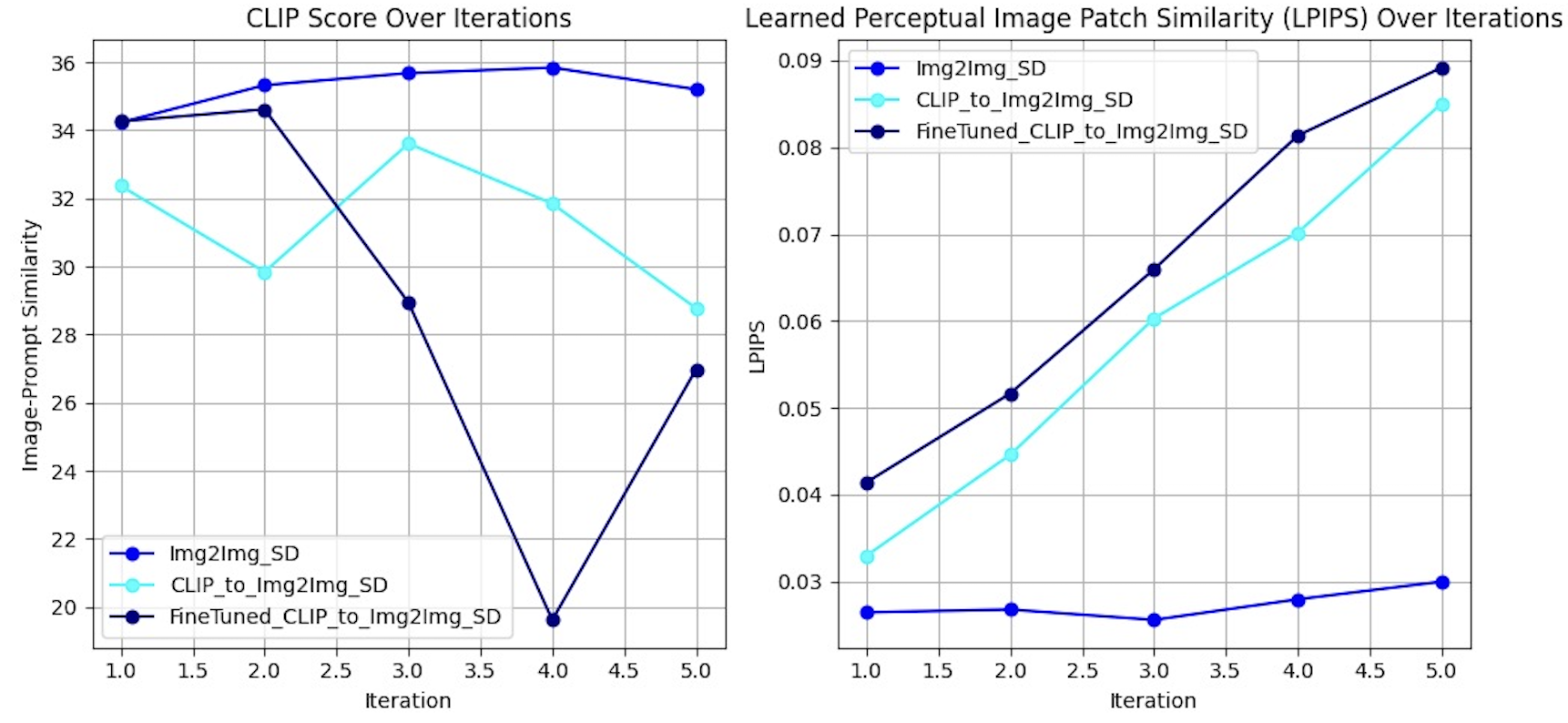}
    \caption{Left: CLIP score over iterative refinement. Right: LPIPS over iterative refinement. Model (1) consistently has the highest CLIP score across all iterations performing 9.17\% better than Model (2) and 1.61\% better than Model (3) and due to lower LPIPS values, closer resemblance between iterations.}
    \label{fig:tests2}
\end{figure}

To evaluate performance across the three models, we tested them using SSIM, PSNR, CLIP score, and LPIPS metrics. Additionally, iterative refinement was performed over five iterations for each model. Figures \ref{fig:tests1} and \ref{fig:tests2} display the results of the performance tests.

\begin{figure}[h!]
    \centering
    \includegraphics[width=0.5\textwidth]{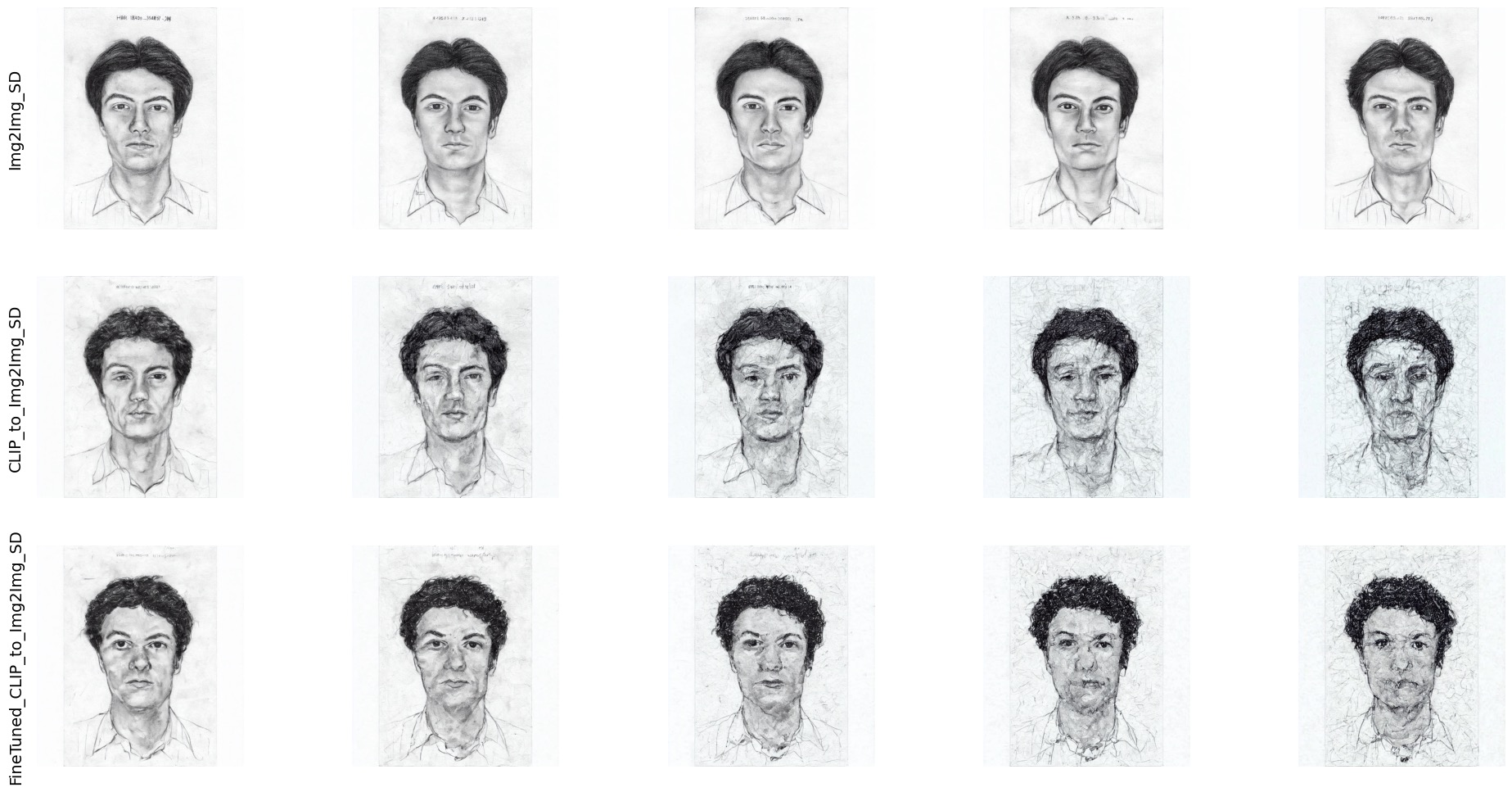}
    \caption{Top: Model (1). Middle: Model (2). Bottom: Model (3). Model (1) has more consistency and visibly less distortion and grain across iterations. Model (3) experiences heavy distortion by the third iteration.}
    \label{fig:sketch_tests}
\end{figure}

Figure \ref{fig:sketch_tests} shows the sketches generated by each model across five iterations. This provides a qualitative comparison of how iterative refinement affects the visual quality of the generated sketches.

\section{Analysis}

As shown in Figures \ref{fig:tests1} and \ref{fig:tests2}, Model (1) outperformed both Model (2) and Model (3) across most metrics.  It achieved the highest SSIM of 0.72, indicating superior alignment with ground truth images, and the best PSNR at 25 dB, reflecting reduced distortion and improved pixel-level clarity. Model (1) also maintained the highest CLIP score throughout iterations, demonstrating strong text-image alignment, and consistently achieved the lowest LPIPS values, indicating closer perceptual resemblance to ground truth images. While Model (3) showed some improvements over Model (2) in CLIP scores and LPIPS, it still trailed behind Model (1) across all metrics.

\section{Related Work}


The baseline for this work, Model (1), is the state-of-the-art Image-to-Image Stable Diffusion model by~\cite{rombach2022latent}, which combines high-quality image generation with computational efficiency. This model serves as the foundation for advancing sketch generation.

Model (2) extends this baseline by integrating the pre-trained CLIP model~\cite{radford2021clip}, leveraging its ability to align text and image features in a shared space. This improves text-to-sketch alignment, outperforming conventional text-to-image methods in semantic accuracy and contextual relevance.

Our novel Model (3) incorporates LoRA-based fine-tuning~\cite{hu2021lora} of CLIP, refining alignment in self- and cross-attention layers. Compared to Nichol et al.'s GLIDE model~\cite{nichol2021glide}, which excels in photorealism but lacks sketch-specific precision, Model (3) offers enhanced alignment and computational efficiency, addressing key limitations in prior approaches.

By combining latent diffusion, multimodal alignment, and efficient fine-tuning, our method sets a new benchmark for text-to-sketch generation.

\section{Conclusion}

This project contributes to the literature by presenting a novel AI-driven approach for police sketch generation, leveraging multimodal inputs and iterative refinement. By integrating LoRA-based fine-tuning with Stable Diffusion, we provide an alternative to traditional manual sketch artistry, addressing key limitations in efficiency and consistency. 

We learned, firstly, the importance of balancing model complexity and performance. Second, data consistency plays a crucial role in achieving reliable outcomes. Third, while Model (1) currently outperforms the other two models, the iterative refinement step demonstrates promise in improving CLIP Scores and LPIPS for Model (3). This suggests that with further optimization, Model (3) has the potential to surpass Model (1) in specific applications. The results indicate that fine-tuning CLIP with LoRA and incorporating iterative feedback mechanisms remain valuable areas for future exploration.


Future work includes expanding the dataset and to investigate the limitations imposed by the 77-token input restriction in CLIP models. This constraint hinders the ability to capture nuanced facial differences, and exploring methods to extend this limit could significantly enhance model performance for this task. Additionally, further exploration of iterative refinement strategies, including the use of masking to focus on specific facial features, could improve the alignment of CLIP embeddings with targeted modifications and achieve better results.

\section{Group Contribution Statement}

Nico Fidalgo built the pipeline for Models (1), (2), and (3). He fine-tuned the CLIP model for Model (3) and conducted the ablation study. He also generated initial sketches with each model as a sanity check.

Katherine Harvey built the dataset for fine-tuning as well as the testing script to evaluate the models' performance across various metrics and sketch generation over iterations. She also incorporated the CLIP model embeddings of the inputs (and additions in the iterative step) into implement into the SD model in Models (2) and (3).

Johnny Ni analyzed model fine-tuning performance, evaluating alternative sketch-generation pipelines, and dataset integrity.

Aaron Contreras built and fine-tuned an alternative Kandinsky image-to-image stable diffusion pipeline to Model (1) with ControlNet.


\begin{thebibliography}{00}

\bibitem{nichol2021glide}
A. Nichol, P. Dhariwal, A. Ramesh, P. Mishkin, B. McGrew, I. Sutskever, and M. Chen, ``GLIDE: Towards Photorealistic Image Generation and Editing with Text-Guided Diffusion Models,'' arXiv, 2021. [Online]. Available: https://arxiv.org/abs/2112.10741. Accessed: Nov. 10, 2024.


\bibitem{radford2021clip}
A. Radford, J. W. Kim, C. Hallacy, A. Ramesh, G. Goh, S. Agarwal, G. Sastry, et al., ``Learning Transferable Visual Models From Natural Language Supervision,'' ICML, pp. 8748--8763, 2021. [Online]. Available: https://arxiv.org/abs/2103.00020. Accessed: Nov. 10, 2024.
\bibitem{hu2021lora}
E. J. Hu, D. Shen, P. Wallis, Z. Allen-Zhu, Y. Li, S. Wang, L. Wang, et al., ``LoRA: Low-Rank Adaptation of Large Language Models,'' NeurIPS, 2021. [Online]. Available: https://arxiv.org/abs/2106.09685. Accessed: Nov. 10, 2024.


\bibitem{b2} Hugging Face, "LoRA: Low-Rank Adaptation of Large Language Models," Hugging Face Blog. [Online]. Available: https://huggingface.co/blog/lora. Accessed: Nov. 10, 2024.
\bibitem{b5} J. Nilsson and T. Akenine-Möller, ``Understanding SSIM,'' NVIDIA. 2020. Available: arXiv, https://doi.org/10.48550/arXiv.2006.13846.

\bibitem{b1}OpenAI, "CLIP (Contrastive Language–Image Pretraining)," OpenAI. [Online]. Available: https://openai.com/index/clip/. Accessed: Nov. 10, 2024.
\bibitem{b3} OpenAI, ``OpenAI CLIP Bit Base Patch32 Model,'' Hugging Face, 2023. [Online]. Available: https://huggingface.co/openai/clip-bit-base-patch32

\bibitem{rombach2022latent}
R. Rombach, A. Blattmann, D. Lorenz, P. Esser, and B. Ommer, ``High-Resolution Image Synthesis with Latent Diffusion Models,'' CVPR, pp. 10684--10694, 2022. [Online]. Available: https://arxiv.org/abs/2112.10752. Accessed: Nov. 10, 2024.

\bibitem{b4} Stability AI, ``Stable Diffusion v1.5 Model,'' Hugging Face, 2022. [Online]. Available: https://huggingface.co/stable-diffusion-v1-5/stable-diffusion-v1-5





\end{thebibliography}
\end{document}